\documentclass{article}

\usepackage[final,nonatbib]{neurips_2020}

\usepackage[utf8]{inputenc} % allow utf-8 input
\usepackage[T1]{fontenc}    % use 8-bit T1 fonts
\usepackage{hyperref}       % hyperlinks
\usepackage{url}            % simple URL typesetting
\usepackage{booktabs}       % professional-quality tables
\usepackage{amsfonts}       % blackboard math symbols
\usepackage{nicefrac}       % compact symbols for 1/2, etc.
\usepackage{microtype}      % microtypography

\usepackage{comment}
\usepackage{xcolor}
\usepackage{multirow}
\usepackage[T1]{fontenc}
\usepackage[font=small,labelfont=bf,tableposition=top]{caption}

\DeclareCaptionLabelFormat{andtable}{#1~#2  \&  \tablename~\thetable}

\usepackage{blindtext}

\usepackage{times}
\usepackage{epsfig}
\usepackage{graphicx}
\usepackage{amsmath}
\usepackage{amssymb}

\usepackage[para,online,flushleft]{threeparttable}

\usepackage{graphics}

\title{Heterogeneous Federated Learning}

\author{
\textbf{Fuxun Yu$^1$, ~~~~~Weishan Zhang$^1$,~~~~~ Zhuwei Qin$^1$,~~~~~ Zirui Xu$^1$} \\ \vspace{2mm}
\textbf{Di Wang$^2$,~~ Chenchen Liu$^3$,~~ Zhi Tian$^1$,~~ Xiang Chen$^1$} \\ 
\textit{$^1$George Mason University, $^2$Microsoft} \\ \vspace{2mm}
\textit{$^3$University of Maryland, Baltimore County} \\ 
\textit{\{fyu2, wzhang23, zqin, zxu21, ztian1, xchen26\}@gmu.edu}\\
\textit{wangdi@microsoft.com, ccliu@umbc.edu}
}

\begin{document}

\maketitle
%\graphicspath{{_fig/}}

\vspace{-3mm}
\begin{abstract}
Federated learning learns from scattered data by fusing collaborative models from local nodes.
	However, due to chaotic information distribution, the model fusion may suffer from structural misalignment with regard to unmatched parameters.
	In this work, we propose a novel federated learning framework to resolve this issue by establishing a firm structure-information alignment across collaborative models.
	Specifically, we design a feature-oriented regulation method (\textit{$\Psi$-Net}) to ensure explicit feature information allocation in different neural network structures.
	Applying this regulating method to collaborative models, matchable structures with similar feature information can be initialized at the very early training stage.
	During the federated learning process under either IID or non-IID scenarios, dedicated collaboration schemes further guarantee ordered information distribution with definite structure matching, so as the comprehensive model alignment.
	Eventually, this framework effectively enhances the federated learning applicability to extensive heterogeneous settings, while providing excellent convergence speed, accuracy, and computation/communication efficiency.
\end{abstract}
\vspace{-2mm}
\section{Introduction}
\label{sec:intro}
\vspace{-3mm}

Federated Learning (FL) has drawn great attention with outstanding collaborative training performance and data privacy supportability~\cite{fl}.
	It is commonly achieved by fusing collaborative nodes' homogeneous neural networks through Federated Averaging (FedAvg), which generates a global model by averaging local parameters with the same coordinates~\cite{fl, fedma}.
	However, FedAvg suffers from non-negligible accuracy drop due to inevitable heterogeneity: highly Non-IID data across nodes may cause huge divergence within the parameters to be averaged and thus significantly skewed results~\cite{non-iid0, iclr}.
	%and therefore significance performance degradation~\cite{niid}.
	%performance drop is mainly caused by averaging diverged parameters, which is called parameter divergence~\cite{niid}.
	%when learning rate is not well tuned or the data distributions are highly non-iid~\cite{niid}.
	%Previous hypotheses are that the performance drop of FedAvg is mainly caused by averaging diverged parameters, which is called parameter divergence~\cite{niid}.

Recent works further elaborate on such heterogeneity from a structural alignment perspective~\cite{rmatching, fusion, peers}.
	Although local neural network models have the homogeneous architecture, the parameters with the same model coordinates might learn different features at different levels, resulting in chaotic parameter distribution across collaborative nodes.
	Therefore, the conventional coordinates-based averaging scheme might fuse unmatched parameters and continuously hurt the global convergence~\cite{fedma, pfnm}.
	%Although local models have the homogeneous architecture, their parameters with the same model coordinates might learn different features, causing the misalignment of structure and  between models.
	%When causing the misalignment of similar parameter structure and information between models~\cite{fedma}.
	%Although local models have the homogeneous structure, their parameters with the same model coordinates might learned different feature information, causing gradually expanding permutation variance~\cite{fedma}.
	%Therefore, the conventional coordinates-based averaging scheme would result in unmatched parameter fusion and continuously hurt the global convergence.
	% Without considering the permutation variance, the conventional averaging scheme would result in unmatched information fusion and continuously hurt the global convergence.
	%
	%Following that, recent works further showed the parameter divergence is one consequence of the permutation-invariance of neural networks~\cite{fedma}.
	%That is, two homogeneous models can have different feature encoding ordering in the structure.
	%Without considering the encoded feature sequences, the coordinate-wise weight averaging could lost learned information and thus hurt the convergence performance~\cite{fedma}.

	Several Federated Learning optimization works have been proposed to leverage this structural alignment property (\textit{i.e.}, ``permutation invariance''~\cite{fedma}), such as Representation Matching~\cite{rmatching}, Bayesian Matching~\cite{bayesian}, FedMA~\cite{fedma}, \textit{etc}.
	%To overcome the permutation variance, several works are proposed recently, such as Representation Matching~\cite{representation}, Bayesian Matching~\cite{PFNM}, FedMA~\cite{fedma}, \textit{etc}.
	%matching-based federated averaging algorithms,
	They conform to a common methodology: after a certain amount of local training, they evaluate parameter similarity across local models and reorganize the permutation, so that approximate parameters could be averaged together.
	Although outperformed conventional schemes like FedAvg, these methods still have certain limitations, such as inaccurate parameter matching, extra computation/communication overhead, limited heterogeneous applicability, and compromised data privacy.

\begin{comment}
\textcolor{purple}{(1)	The non-interpretable matching metrics cannot accurately reflect the similarity of encoded features. Most matching metrics usually evaluated the weight similarity of feature similarity by MSE loss to match the averaging units, while small MSE loss does not necessarily indicate similar features;}

\textcolor{purple}{(2)	Most matching algorithms often assume different local models have mostly matchable units. While when facing data heterogeneity, especially highly-skewed non-iid data, different local models can carry distinct features and a forced matching can cause catastrophic information loss;}

\textcolor{purple}{(3)	The post-matching algorithm incurs heavy extra communication overhead due to uploading weight/feature for similarity calculation. Besides, using feature-based matching metrics can further require sharing private data to get the features, which compromises the data privacy.}
\end{comment}

\begin{figure*}[t]
	\begin{center}
	\vspace{-5mm}
	\includegraphics[width=5.5in]{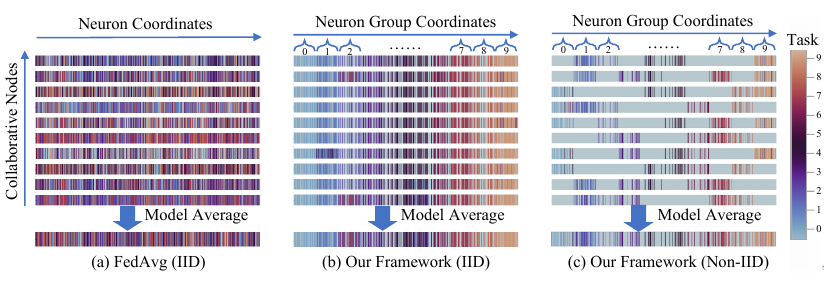}
	\vspace{-7mm}
		\caption{\small{The feature encoding comparison between FedAvg and our proposed framework. 
		The color of each neuron is determined by its top response class to indicate its learned feature, as defined in Eq.~\ref{eq:2}. 
		\textbf{(a)} The original FedAvg with chaotic feature encoding can suffer from feature averaging conflicts among different nodes. 
		\textbf{(b) \& (c)} In contrast, our framework enforces structurally-aligned feature encoding and alleviates such averaging conflicts in both IID and non-IID cases.
		Experiments are conducted with 10 collaborative nodes (VGG9 on CIFAR10). IID: Each node has data of 10 classes. Non-IID: Each node has data of 5 varied classes.}}
	\label{fig:motivation}
	\vspace{-6mm}
	\end{center}
\end{figure*}

To tackle these limitations, we propose a novel FL framework to resolve this issue by establishing not only structure but also information alignment across collaborative models.
	Specifically, we design a feature-oriented regulation method (\textit{$\Psi$-Net}), which can effectively identify local models' structural hierarchies and grouped components that are associated with particular classes and corresponding features.
	Based on such explicit feature information allocation, matchable structures with similar feature information can be initialized at the very early training stage, and further guarantees ordered information distribution with definite structure matching along the whole training process.
	% so as the comprehensive model alignment for FL performance enhancement.
	Fig.~\ref{fig:motivation} provides a set of intuitive comparisons for parameter matching across collaborative models.
	With the proposed feature allocated regulation, our model's encoded feature distribution conform to structural alignments with corresponding classes, while regular FedAvg's model fusion suffers from significant parameter mismatching and therefore expected performance degradation.
	
We conduct extensive experiments on general datasets (CIFAR10, 100) with two model architectures to regulate (VGG16 and MobileNet).
	Our work demonstrates significant improvement in both convergence speed and accuracy, outperforming previous state-of-the-art works by large margins (+2.5\%$\sim$4.6\%) with significantly less computation and communication cost.
	Especially under highly-skewed non-IID scenarios, our work still performs effective and robust model alignment to facilitate the convergence while most previous methods cannot.
	To our best knowledge, this is the very first work that solves the FL alignment problem directly from a model regulation perspective, which significantly improves the FL applicability in various heterogeneous settings.

\normalsize
\vspace{-2mm}
\section{Background and Motivation}
\label{sec:2}
\vspace{-3mm}

\paragraph{Neural Network Permutation Invariance.}
\begin{comment}
	%With its great communication efficiency and deployment simplicity, FedAvg has been the most popular collaborative training algorithm in the past few years.
	%As aforementioned, recent works have shown that due to the permutation invariance of neural networks, the coordinate-wise weight averaging can
\end{comment}
From a permutation perspective, the model parameters (\textit{e.g.}, neuron weight matrix $\Omega_i$ in the $i$\textit{th} node's model) can be decoupled as $\omega_i\Pi_i$, where $\omega_i$ defines parameter values and $\Pi_i$ defines the alignment sequence.
	For $N$ collaborative training nodes with the homogeneous neural network architecture, the model permutation invariance can be formulated as:
\small
\begin{equation}
	F(X) = (\omega_0 \Pi_0)\Pi_0^TX = \cdots = (\omega_i \Pi_i)\Pi_i^TX = \cdots = (\omega_N \Pi_N)\Pi_N^TX.
	\hspace{5mm}
	%N~\textrm{collaborative~training~nodes}.
	\label{eq:1}
\end{equation}
\normalsize
	Although the local models across collaborative nodes can achieve the same global function $F$ on the input set $X$, the local models' $\omega$ and $\Pi$ can vary from model to model~\cite{fedma, pfnm}.
	%structural alignment can be formulated by model parameters (\textit{e.g.}, neuron weight matrix $W_i$ in the $i$\textit{th} node's model) along with the corresponding permutation matrix $\Pi_i$, while representing the global function $F$ on individual input $X$:
	In such circumstances, their encoded feature distribution can vary due to the permutation matrix $\Pi$, causing considerable parameter mismatching ($\omega_i\Pi_i \neq \omega_j\Pi_j$) and thus collaborative training performance degradation.
\begin{comment}
For $N$ collaborative training nodes with the homogeneous neural network architecture, the model structural alignment can be formulated by model parameters (\textit{e.g.}, neuron weight matrix $W_i$ in the $i$\textit{th} node's model) along with the corresponding permutation matrix $\Pi_i$, while representing the global function $F$ on individual input $X$:
	%Specifically, for a homogeneous structure (convolutional or fully-connected layer), the optimal neuron weights $W_i$ can be of varied permuting orders $\Pi_i$, while represent the similar function F on input X:
\small
\begin{equation}
	F(X) = (W_0 \Pi_0)X = \cdots = (W_i \Pi_i)X = \cdots = (W_N \Pi_N)X.
	\hspace{5mm}
	%N~\textrm{collaborative~training~nodes}.
	\label{eq:1}
\end{equation}
\normalsize
Ideally, a fixed $\Pi$ represents the default coordinates of individual neuron parameter.
	However, during practical training, the permutation matrix $\Pi$ will vary to $\Pi'$ from model to model, causing considerable parameter mismatch to the averaging target $W_i\Pi_i' \neq W_j\Pi_j$, and therefore considerable performance degradation under regular FedAvg scheme.
	%where $\Pi_i$ is a permutation matrix in the $i_{th}$ node's model, which can adjust the parameter alignment order with other nodes.
	%the models in different training nodes.
	%Without enforcing the same ordering $\Pi_i \neq \Pi_j$ on different training nodes, FedAvg conducts agnostic coordinate-wise averaging $W_i + W_j$.
	%Therefore, this can lead to unmatched weight averaging, i.e., the parameter divergence issue.
\end{comment}

\vspace{-2.5mm}
\paragraph{Matched Averaging with Permutation Adjustment.}
To achieve FL model alignment, previous methods mainly improve FedAvg by iteratively matching approximate parameters with permutation adjustment before each global averaging cycle.
	Their common approach is to first identify approximate parameters by evaluating certain parameter characteristics (\textit{e.g.}, weight value or activation degree) across collaborative nodes based on a similarity metric (mostly \textit{MSE} loss).
	Then, they will search an adjustment matrix $\Pi_i^{trans}$ for reordering to achieve the minimum divergence with the averaging target: $\omega_i\Pi_i\Pi^{trans} \approx \omega_j\Pi_j$.
	Specifically, such a problem can be resolved by optimization algorithms similar to solving the Wasserstein Barycenter problem~\cite{opt_trans}.
	%the parameter divergence can be quantitatively evaluated by weight value and activation loss function can be embodied by weight activation. And xxxxx can be achieved by x algorithm (e.g. warycenter)
	%re-align the chaotic weight order, i.e., $\Pi_i\Pi_{adj} = \Pi_j$.
\begin{comment}
	For example, Representation Matching~\cite{representation}, Bayesian Matching~\cite{PFNM}, FedMA~\cite{fedma}, etc.
	Most of these methods conform to the following process: After certain local training epochs, they optimize an weight alignment loss based on weight/feature similarity (mostly MSE loss) to derive the optimal $\Pi_{trans}$ and then re-permute/match their indexes of trained weights for averaging.

However, the post-alignment accurateness is limited by the weight/feature similarity metrics, which is usually non-interpretable and hard to evaluate quantitatively by MSE loss~\cite{fedma}.
	Meanwhile, the feature similarity measurement also requires sharing private data, which compromises the data privacy and introduces extra matching communication overhead~\cite{sysml}.
	Moreover, in FL with highly-skewed task non-iid data distributions, the encoded features of different models are intrinsically non-matched. In such cases, a forced-matching can lead to the feature information loss on both sides.
\end{comment}

\vspace{-2.5mm}
\paragraph{Limitations of the State of the Art.}
Although the current parameter matching works alleviated the FL model alignment issue, they still suffer from certain limitations:
% \\\-\hspace{5mm}

\vspace{-1.5mm}
	\noindent (1) \textit{Inaccurate Parameter Matching}: The state of the art's (SOTA) structural alignment accurateness is limited by the quantitative parameter similarity evaluation.
	The parameters with the least \textit{MSE} difference in weight/activation may carry distinct feature information, and therefore the forced matching without qualitative verification can cause catastrophic information loss~\cite{rethink,bmvc};

\vspace{-1.5mm}
	(2) \textit{Heterogeneous Applicability}: SOTA assumes different local models have mostly matchable parameters ($\omega_i \approx \omega_j$).
	While the practical FL heterogeneity may involve highly Non-IID local data and even asymmetric learning tasks, leading huge model alignment complexity~\cite{rmatching,iclr};

\vspace{-1.5mm}
	(3) \textit{Performance Overhead}: The post-training parameter matching incurs extra computation load for parameter similarity evaluation and permutation adjustment optimization.
	Meanwhile, parameter similarity evaluation across nodes also introduces considerable communication overhead;

\vspace{-1.5mm}
	(4) \textit{Data Privacy}: Moreover, the activation-based similarity evaluation requires data sharing between different nodes, thus the data privacy can be compromised~\cite{opt_trans}.

\vspace{-2.5mm}
\paragraph{Design Motivation: Feature-Level Alignment.}
	We expect to relieve the above problems from two aspects:
	On one hand, beyond feature-agnostic parameter similarity evaluation, we would further interpret the encoded feature information in neural network components and promote the parameter similarity assessment to the feature level. %, so as to achieve more accurate parameter matching in both structure-wise and information-wise.
On the other hand, we would design a feature-oriented model regulation method to ensure explicit feature allocation in specific model structures.
	Rather than post-training alignment, such a regulation method adjusts local model architecture according to their data and task distribution at the very early training stage, and continuously maintain not only structure but also information alignment for better FL performance.
	%Eventually, this framework is expected to effectively enhance the FL applicability to extensive heterogeneous settings with optimal performance.

Motivated by these two aspects, our work begins with the interpretation of encoded feature information in neural networks.
	Specifically, we adopt a neuron's class response preference to indicate its learned feature~\cite{bmvc,vis1,vis2, vis4}:
	Starting with single fully-connected layer for classification tasks, such a preference can be measured by observing the neuron's activation response $A(x_c)$ on inputs $x$ from $C$ classes, as well as its gradients {\small $\partial Z_c / \partial A(x_c)$} (or weights $w_c$ equally) towards a class $c$'s prediction confidence $Z_c$.
	Combining these two factors and further generalizing to a multi-layer convolutional neural network, a neuron's learned feature can be indicated by its class preference vector:
\small
\vspace{-1mm}
\begin{equation}
	P = [p_1,\ldots, p_c,\ldots,p_C], \hspace{5mm}\text{where}~p_{c} = \sum_b^B A(x_{c,b}) * \frac{\partial Z_c}{\partial A(x_{c,b})},
	\label{eq:2}
\end{equation}
\normalsize
where $A~(x_{c,b})$ denotes activations and {\small $\partial Z_c/\partial A(x_{c,b})$} denotes gradients from class $c$'s confidence, both of which are averaged on $B$ batch trials.
	The effectiveness of such an interpretation approach is illustrated in Fig.~\ref{fig:motivation}, which demonstrates the parameter/feature divergence from a functionality perspective and highlights the advantages of our feature-level structural information alignment.

\vspace{-2.5mm}
\paragraph{Our Work: Feature-Allocated Model Regulation and Collaboration.}
Based on the feature interpretation, our major work focuses on the second motivation aspect.
	Specifically, we propose a novel federated learning framework including feature-regulated model design and a set of robust collaboration policies under various heterogeneous settings.
	Fig.~\ref{fig:shuffle} presents the framework overview.
	Leveraging feature distribution exploration with Eq.~\ref{eq:3}, we designed a \textit{$\Psi$-Net} model structure that consists of shared layers and grouped layers, which are expected to share fundamental features with all learning classes and isolate class-specific features within exclusive structures, respectively.
	By regulating existing model architectures into such a \textit{$\Psi$-Net} structure, we can maintain features' consistent structural allocation in different training stages, and thereby establishing the basis for the feature level model alignment.
	The collaboration policies can further specify the optimal \textit{$\Psi$-Net} configuration and particular model fusion schemes for various FL scenarios.
	Not only capable for regular IID and non-IID data distribution, our framework can even support extreme heterogeneous cases, where certain classes might be unlearnable on particular collaborative nodes.
	Eventually, this framework is expected to effectively enhance the FL applicability to extensive heterogeneous settings, while providing excellent convergence speed, accuracy, and computation/communication efficiency.
\begin{comment}
	Specifically, we propose a \textit{$\Psi$-Net} model structure regulation by utilizing group convolution to regulate the feature encoding in each groups (Fig.~\ref{fig:shuffle}(a)).
	With the group-separable structure design, local trained models can achieve structural feature encoding, as well as providing an explicit structure-feature mapping relationship, i.e., feature allocation.
	Then for the model collaboration, we propose a corresponding group-matched collaboration policy to conduct weight averaging only within the feature-shared groups(Fig.~\ref{fig:shuffle}(b)).
	As these groups learn mostly similar task features, such averaging avoids the previous feature-level divergence caused by non-matched feature conflicts.
	Finally, the global model with full model structure can be attained by structurally averaging all the participated models(Fig.~\ref{fig:shuffle}(c)).
	By our feature allocated model structure design and collaboration policy, our framework could consistently outperform state-of-the-art methods in both IID and non-IID scenarios, providing higher convergence speed and accuracy.
\end{comment}

\vspace{-3mm}
\section{\textit{$\Psi$-Net} Structure for Feature Allocation}
\label{sec:3}

\vspace{-3mm}
\subsection{Feature Allocation Definition}
\vspace{-3mm}

For a given neural network, the goal of feature allocation can be defined by a structure-feature mapping relationship ($M$) between the model meta-structure sets ($S$) and task sets ($T$):
\small
\vspace{0.5mm}
\begin{equation}
	M = \{s^t_i : t ~~\text{or}~ <t_{0}, ..., t_{n}> \}, \\
	~~\text{s.t.}~~\forall i \neq j, ~~ M(s_i) \cap M(s_j) = \emptyset ,
	\label{eq:3}
\end{equation}
% \vspace{0.5mm}
\normalsize
where $s^t_i$ indicates one meta-structure (\textit{e.g.,} one set of neurons in the neural network) mapped to one (or one set of) task feature $t$, and $M(s^{t1}) \cap M(s^{t2}) = \emptyset$ enforces that any individual task feature can be assigned to only one meta-architecture.
%
% The goal of \textit{$\Psi$-Net} structure design to enforce a pre-aligned structural feature allocation ordering $\Pi_{pre}$ for all nodes before the training process starts:
% \small
% \begin{equation}
% 	F(X) = (W_0 \Pi_{pre}) ~\Pi_{pre}^T X = \cdots (W_i \Pi_{pre}) ~\Pi_{pre}^T X \cdots= (W_N \Pi_{pre}) ~\Pi_{pre}^T X.
% 	\label{eq:3}
% \end{equation}
% \normalsize
% Thus, different nodes could form similar feature encoding during local training, and facilitates the matched weight averaging $W_i\Pi_{pre} + W_j\Pi_{pre}$ in the following collaboration, as shown in Fig.~\ref{fig:motivation} (b).

In general, due to the insufficient interpretability, it's hard to either find the meta-structure $s_i$ or interpret the feature $t$ in a convolutional neural network, not to mention manipulating its feature encoding mapping.
But interestingly, the AlexNet authors found that AlexNet learns distinct shape-oriented and color-oriented features in its two separated convolution groups~\cite{alexnet}, so did ShuffleNet~\cite{shufflenet} with their findings that features become biased among different convolutional groups.
	The two networks both utilize the group convolution structure. 
	Although for different purposes, they demonstrate the similar feature regulation effects.
Following such observation, our hypothesis is in the densely-connected convolutional structures, the training gradients carrying task features can usually backpropagate through the entire network, leading to random feature encoding orders.
	In contrast, the group convolution separates different groups and build implicit boundaries between them, thus preventing gradients flowing interleaved and achieving certain feature regulation effects.

From this perspective, we propose our \textit{$\Psi$-net} regulation utilizing grouped convolutional and linear layers to build a meta-structure sets $S$.
	Meanwhile, by allocating the class logit $t$ to the structure group $s_i$, we assume that the feature of task $t$ is also allocated to the structure, $s_i \rightarrow s^t_i$. 
	In this way, the meta-structure $s^t_i$ always acts as the feature anchor for task feature $t$ regardless of on which nodes $i$, thus ensuring the feature alignment within structure $s^t$ among all nodes:
\small
\vspace{0.5mm}
\begin{equation}
	\Omega_t \leftarrow s^t_i \approx ... s^t_j \approx ... s^t_N, ~~~ \forall ~i, j \in N.
	\label{eq:4}
\end{equation}
% \vspace{0.5mm}
\normalsize
Meanwhile, even though different nodes can have varied structure orders, their local structure $s^t_i$ can still be explictly mapped with the $\Omega_t$, which is the $t$\textit{-th} meta-structure in the global model structure $\Omega$. 
	With such a feature alignment property, we are then able to enforce the structural feature alignment and achieve even heterogeneous model collaboration, as we will show later.

\begin{figure*}[t]
	\begin{center}
	\vspace{-5mm}
	\includegraphics[width=5.5in]{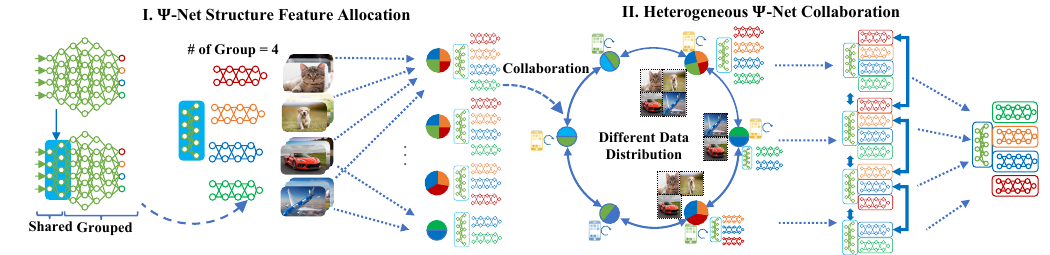}
	\vspace{-4mm}
	\caption{Our proposed heterogeneous learning framework includes two major steps: \textbf{(I)} We utilize group-convolution based \textit{$\Psi$-net} structure as a regulation for feature allocation. \textbf{(II)} Then we propose the heterogeneous \textit{$\Psi$-net} FL policy by grouped-shared parameter averaging to handle both IID and non-IID data distributions.}
	\label{fig:shuffle}
	\vspace{-6mm}
	\end{center}
\end{figure*}

\vspace{-3mm}
\subsection{\textit{$\Psi$-Net} Structure Implementation} 
\vspace{-3mm}
As shown in Fig.~\ref{fig:shuffle} (a), \textit{$\Psi$-Net} structure regulation is composed of two major parts to achieve grouped-based feature allocation: Non-grouped shared layers (lower convolutional layers) and grouped layers (higher convolutional and fully-connected layers).

\vspace{-3mm}
\paragraph{Shared Layers.} The design of shared layers is due to the shallow layers learn mostly basic features that are useful for most tasks and show less feature divergence. 
	In such cases, grouping these layers can prevent neurons in one group from receiving gradients from other groups, leading to sub-optimal learning performance.
	Therefore, the shallow layers are reserved shared without grouping.

To determine an appropriate number of shared layers to reserve, we evaluate the feature divergence of layer $l$ based on the total variance ($TV$) of neurons' feature encoding:
\small
% \vspace{-1mm}
\begin{equation}
	TV_l = \sum\nolimits_i^L \frac{1}{L}|| P_{l,i} - E(P_{l,i})||_{2}, ~~ \text{where} ~~ P_{l,i} = [~p_{l,1}, ..., p_{l,c}, ..., p_{l,N}]/\sum\nolimits_j^N{p_{l,j}},
	\label{eq:5}
\end{equation}
% \vspace{-1mm}
\normalsize
where we use the neuron's class preference vector $P_{l,i}$ in Eq.~\ref{eq:2} as an indication of its learned feature. 

From shallow to deep layers, the layer-wise feature divergence can surge to high points in some medium levels. 
	We thus evaluate the feature divergence and empirically determine a sharing depth to start the grouped layer construction.
	In later experiments, we will evaluate our depth selection and show that our \textit{$\Psi$-net} structure is fairly robust to the depth hyper-parameter selection.

\vspace{-3mm}
\paragraph{Group Convolutional Layers.} For the deeper convolutional layers, the encoded class feature are more diverged and thus easier to conflict with each other during weight averaging.
	Therefore, we utilize group convolution and construct separable group convolutions.
	In each group, the convolutional neuron can only receive/output the feature channels within the current group.
	Thus, the backpropagated gradients carrying feature information can only flow through the individual group, avoiding the feature interleaving between different groups.

Considering the practical training dataset complexity, the number of groups cannot be always ``one-to-one'' matched to the training tasks.
	Therefore, on big datasets (\textit{e.g.,} CIFAR100 with 100 classes), we also supports ``one-to-many'' mapping, \textit{i.e.}, multiple class logits mapped to one convolutional group, so as to achieve high scalability but without sacrificing the feature regulation performance too much.
	As we will show later, both one-to-one mapping on small datasets (CIFAR10) and one-to-many mapping on mediate datasets (CIFAR100) can achieve as good structural feature regulation benefits.

% As the group convolution constructs the implicit feature boundary, the next step is to enforce the feature sources for each group so as to achieve the desired feature allocation. 
% 	To do so, we thus adapt the fully-connected (FC) layers into grouped linear layers.

\vspace{-3mm}
\paragraph{Grouped Fully-Connected Layers.} For the fully-connected (FC) layers, the previous FC layers often connect all input convolutional neurons to the class logits, and thus cannot regulating which convolutional group learns the corresponding task features. 
	In \textit{$\Psi$-net}, we also decouple the original FC layer into groups, \textit{i.e.}, multiple linear layers with same number as the convolutional groups.
	One linear layer connects the class logit(s) with the mapped convolutional groups only, thus enforcing the gradients flowing backwards to the mapped group without any leakage. 
	Multi-layer FC layers can be decoupled similarly across layers, which we omit for simplicity.

\vspace{-3mm}
\paragraph{Group Normalization Layers.}
Previous works have shown that batch normalization layers can influence the FL performance as different local models tend to collect non-consistent batch mean and variance statistics (especially in non-IID cases), thus degrading the performance~\cite{iclr}. 
Our \textit{$\Psi$-net} structure design, by enforcing the similar feature encoding, alleviates the batch statistics divergence within each group.
	Therefore, we propose to incorporate the Group Normalization (GN) layer~\cite{group_norm} into our \textit{$\Psi$-net} structure to improve the model convergence performance. 
	Specifically, we add the GN layer after each grouped convolutional layer, both with the same number of groups to ensure the batch statistic normalization is conducted within each group.
	For the shared layers, we can regard it as a layer with group number equal to 1, which deduces to the normal batch normalization.
	In the later evaluation, we will demonstrate the effectiveness of GN in our \textit{$\Psi$-net} structure.

\vspace{-0.5mm}
Overall, by our \textit{$\Psi$-net} structure, we can isolate different convolutional groups, so as the task features between them.
	Each conv/fc group only learns the specified task features from the assigned class logits by gradient back-propagation, enabling a pre-aligned feature allocation during local training.
	Also, \textit{$\Psi$-net} regulation is highly configurable and supports dynamic structure-feature mapping, which provides high flexibility in handling training datasets with different complexity.
	% The \textit{$\Psi$-net} structure regulation is also applicable for most CNN structures like VGG or MobileNet, which can greatly improve the FL convergence speed and accuracy over the original structure, as we will show later.

%\input{_txt/4_stru}
\vspace{-2mm}
\section{\textit{$\Psi$-Net} Collaboration for Heterogeneous Federated Learning}
\vspace{-3mm}

In this section, we present the particular FL collaboration framework based on the \textit{$\Psi$-Net} structure regulation through different training stages.
	To demonstrate the framework's applicability for various heterogeneous settings, we target both IID and non-IID data distribution scenarios.
	Especially, we also take an extreme non-IID case into consideration: certain classes' data volume might be too small or missing on particular nodes, resulting in asymmetric learning tasks across local models.

\begin{comment}
Help achieve efficient and effective federated collaboration in both IID and non-IID cases.
	In IID cases, the local and global models share the homogeneous \textit{$\Psi$-net} structure with structural feature allocation.
	Especially in non-IID scenarios, our local \textit{$\Psi$-net} structure can be adapted to form a partial-structure-feature mapping to conduct more efficient local training.
	Finally, we also propose and integrate the novel group-matched model averaging and global model collection policy to accomplish our heterogeneous framework.
\end{comment}

\vspace{-2.5mm}
\paragraph{Global \& Local Model Initialization.}
Regarding the heterogeneity across collaborative nodes, the \textit{$\Psi$-Net} based global and local models might be initialized differently for training scenario adaptation.

\vspace{-1mm}
\textit{Global Model:} Given a global task with multiple learning classes and a target neural network architecture to transform (\textit{e.g.}, VGG16), the proposed regulation method will first identify the depth boundary between shared and grouped layers based on Eq.~\ref{eq:5}.
	% This is done by attaining a pre-trained model (original structure) with small pre-training overhead ($\sim$50 epochs) to get the layer divergence statistics.
Regardless of IID or non-IID scenarios, the number of separated layer groups will be determined with the mapping schemes described in Sec. 3 to guarantee the expected granularity and scalability with different learning class volumes.
	%and exshd layers are selected based the with high feature divergence to support effective structure-feature mapping.
\begin{comment}
\vspace{-1.3mm}
Also, we set the number of groups depending on the global task set complexity. 
	Here the tradeoff is that, larger number of groups provide more fine-grained structure-feature mapping (e.g., one-to-one mapping), but can suffer from insufficient group capacity if the number of neurons in each group is too small.
	As we will evaluate later, our framework is robust to both hyper-parameters' selection (with optimal performance in large ranges of depth and group number selection).
\end{comment}
	% will cover global model is initialized to  is the grouped \textit{$\Psi$-net} model based on the global task sets and the structure-feature mapping relationship in both IID and non-IID cases.
	% As aforementioned, the feature-structure mapping could be determined as one-to-one or one-to-many depending on the global training set complexity.

\vspace{-0.3mm}
\textit{Local Model:} In IID and general non-IID data distribution cases, local models share the homogeneous structure with the global model.
	While in the extreme non-IID cases, where particular nodes are unable to learn certain classes, heterogeneous models will be initialized for the asymmetric learning tasks.
	From the structure-information alignment perspective, the structure groups corresponding to unlearnable classes will be trimmed, as they cannot learn useful information to contribute.
	
\vspace{-1.3mm}
Such a trimming process is implemented differently following the mapping scheme adopted by the global model:
	With the ``one-to-one'' mapping scheme, any structure group without local task data being present would be trimmed.
	While with ``one-to-many'', the trimming of one structure group is conducted only if all the mapped tasks $M(s_c)$ are not present in node $i$'s local data $LD(i)$:
\small
\begin{equation}
	%\omega_{shared}^{new} = \frac{1}{M} \sum_{i = 1}^{M} \omega_{shared}^i
	s_c = \textbf{1}, ~~\text{iff}~~ M(s_c) \cap LD(i) = \emptyset~, ~~\text{otherwise} ~~\textbf{0},
	\label{eq:sharelayer}
	% \vspace{-1mm}
\end{equation}
\normalsize
where $M(s_c)$ is the task set mapped to the current conv/fc group. 
	Better rules may be determined considering the amount of data or abandoning isolated classes, which we leave as future work.

\vspace{-2.5mm}
\paragraph{Local Model Training.}
\begin{comment}
Despite the potential heterogeneous model structures, our local model training stage still minimizes training loss on each node's respective local data:
\small
\begin{equation}
	\min L[F_k(X_k), Y_k], \quad \forall F_k(x) \in \{F_1(x), \dots, F_M(x)\}
	\label{eq:sgd}
	% \vspace{-1mm}
\end{equation}
\normalsize
where $L(.)$ denotes the training cross-entropy loss, $F_k(.)$ denotes the local model owned by node $k$ and $[X_k, Y_k]$ denotes the non-iid local dataset on node $k$.
	$\{F_1(x), \dots, F_M(x)\}$ is the set of all participating local models , where different models $F_i(x)$ can be of varied structure.
\end{comment}
Local training is conducted epoch-wise as regular FL schemes.
	However, different from SOTA, the structure-information alignment is continuously guaranteed without considerable overheads, such as parameter sharing, evaluation, and post-training alignment.
	Moreover, the nodes with asymmetric tasks further trim local models, leading to less computation cost.

\vspace{-2.5mm}
\paragraph{Aligned Model Averaging.}
The averaging will be conducted periodically every $e$ epochs.\vspace{0.5mm}
	%When a specific amount of local training epochs is finished, collaboration is conducted before next training round to preserve global consensus.
	%According to our initialization method, each local sub-model can be spitted into shared layers and separable grouped layers.

\vspace{-1mm}
\textit{Shared layers} will be synchronized by local models from all $N$ collaborative nodes.
	As they extract fundamental features with less class-specific divergence, an global averaging can be directly applied:
\small
\begin{equation}
	%\omega_{shared}^{new} = \frac{1}{M} \sum_{i = 1}^{M} \omega_{shared}^i
	\Omega_{shared} = Avg(\omega_{shared}^{i}),~\hspace{5mm} i \in \{1,\ldots, N\},
	\label{eq:sharelayer}
	% \vspace{-1mm}
\end{equation}
\normalsize
where $\omega_{shared}^i$ denotes the weights of the shared layers from the $i$\textit{th} node's local model.
	Moreover, such uniformed shared layers also benefit for global consensus under heterogeneous scenarios.
	% $\mathbf{M}$ denotes the set of nodes in our FL system, $E(.)$ denotes the average operator.
	% The reason for taking an global average is that shared layers extract fundamental features which show less task divergence.
	% Also, a unified fundamental feature extractor is important for the consensus of functionality modules.

\vspace{-1mm}
\textit{Grouped layers} conduct model averaging only within matched groups, which are supposed to have the same primary learning class to alleviate potential feature conflicts:
\small
\begin{equation}
	\Omega_{c} = Avg(\omega_c^{\mathbf{I}_c}),~\hspace{5mm} \text{where}~\hspace{2mm} \mathbf{I}_c = \{i,~\text{s.t.}~c \in M(i)\},
	\label{eq:localagg}
\end{equation}
\normalsize
where $\Omega_{c}$ denotes the averaged weights of the grouped layer structures for the $c$\textit{th} learning class,
	$\mathbf{I}_c$ indicates the indices of collaborative nodes where class $c$ presents in their local data sets.
	Ideally, all collaborative nodes should be included in $\mathbf{I}_c$.
	However, when asymmetric learning tasks occur, $\mathbf{I}_c$ will exclude certain unsupportive local models for class $c$.
	% Meanwhile, the averaging is also conducted among the same structure $\omega_c$, a.

\vspace{-1mm}
\textit{Normalization layers}, including batch and group normalization layer, are averaged in a similar way:
	For batch normalization in shared layers, we average the ``\textit{shift}'' and ``\textit{scale}'' parameters, as well as the the ``\textit{running\_mean}'' and ``\textit{running\_variance}'' statistics for each layer.
	For group normalization in the grouped layers, only the ``\textit{shift}'' and ``\textit{scale}'' parameters are averaged within matched groups, as the other two are subject to the batch change.
	% (as GN calculates running\_mean and running\_varaince only based on the current batch).

\vspace{-2.5mm}
\paragraph{{Heterogeneous Model Collection.}}
% When each sub-model is well-trained on local, their CNN modules should become stationary and proficient to local tasks.
% Considering the impact of within-group parameter sharing, task-identical CNN modules across neighbor nodes should be both converged and achieve sub-consensus within the same task group:

% \vspace{-2mm}
% \scriptsize
% \begin{equation}
% \begin{split}
% 	\min L[ F_i(X_i), Y_i], \quad &\forall F_i(x) \in \{F_1(x), \dots, F_M(x)\}
% 	\\ \text{s.t. } F_i(x)\cong F_j(x), \quad \forall i,& j\in \{1, \dots, M\} \text{ s.t. } C_{i,j} \neq 0
% 	%X_i\cap X_j \neq \varnothing
% 	\label{eq:locconv}
% 	\vspace{-1mm}
% \end{split}
% \end{equation}
% \normalsize
% where $[X_i, Y_i]$ denotes the local dataset with labels on the node $i$, $\{F_1(x), \dots, F_N(x)\}$ denotes the union set of local sub-models in our framework, ``$\cong$" denotes the consensus on task-identical CNN modules and $C_{i,j}\neq 0 $ means that connection exists between node $i$ and $j$.

% This formulation describes the condition of all sub-models when the whole system is well-trained.
% The unified task-identical CNN modules should be proficient to work with all tasks across collaborating neighbor nodes.
% For a highly connected learning system, CNN modules for each task can reach global consensus and thus can form a effective target model.
Taking the aforementioned extreme non-IID cases into consideration, the final global model can be considered as a collection of heterogeneous local structures:

\vspace{-3mm}
\small
\begin{equation}
	\Omega^{Global} = \{\Omega_{shared},~\Omega_{1},\ldots,~\Omega_{C}\}.
	\label{eq:exm_glob}
\end{equation}
\normalsize
Overall, along with the feature allocated structure regulation, our framework can avoid the collaborative conflicts and achieve continuously feature alignment regulation during different FL stages.
	In next section, we conduct systematical evaluation of our framework.
\vspace{-2mm}
\section{Experiment}
\vspace{-3mm}

\paragraph{Experiment Setup.} 
We evaluate our proposed framework with image classification tasks on CIFAR10 and CIFAR100. 
Three neural network architecture, namely VGG9, VGG16, and MobileNetv1, are adopted to demonstrate the generality of our structure regulation method.
For local data distributions, we consider both IID and non-IID scenarios, where the non-IID can have only partial classes of the dataset.
State-of-the-art works including FedMA~\cite{fedma} and FedProx~\cite{fedprox} are compared to demonstrate the efficiency and effectiveness of our proposed framework.
Code will be released.

\vspace{-4mm}
\subsection{Convergence Performance Improvement Evaluation}
\vspace{-3mm}
In this section, we first demonstrate our framework's advanced performance in terms of convergence accuracy and speed.
	As aforementioned, we adopt the ``one-to-one'' mapping scheme to generate 10 structure groups with \textit{$\Phi$-Net}  on CIFAR10 for fine-grained feature allocation. 
	On CIFAR100, we also utilize the 10-group structure with the ``one-to-many'' mapping in order to achieve the required scalability with increased dataset complexity.

\vspace{-2mm}
\paragraph{Accuracy Improvement (Non-IID to IID).}
We demonstrate our framework's full-spectrum accuracy improvement in Table.~\ref{table:full_spectrum}.
	The experimental setting $N * C$ indicates there are $N$ collaborative nodes, and each node has $C$ classes present in the local data distribution.
	The global dataset is split randomly and equally onto different nodes without any duplication.
The results reveal that our framework consistently outperforms the baseline -- FedAvg by \textbf{+1\%$\sim$4\%} accuracy on VGG9. 
	FedAvg on MobileNetV1 seems to suffer more from the highly-skewed non-IID data distribution. 
	As a result, our framework could achieve \textbf{+6\%$\sim$19\%} accuracy on MobileNetV1.

\vspace{-3mm}
\paragraph{Accuracy Improvement (Node Scalability).}
Similar experiments are conducted to evaluate our framework's generality with node scaling.
	Specifically, we scale the number of collaborative nodes from 10 to 100 with the non-IID data distribution (each node only have 5 classes present in the local data distribution).
	Without loss of generality, our framework provides consistently better performance ranging from \textbf{+2\%$\sim$4\%} on VGG9 and \textbf{+5\%$\sim$11\%} on MobileNetV1.

\vspace{-0.5mm}
The underlying reasons of our accuracy improvement is due to the structurally aligned feature distribution across local models, as demonstrated in Fig.~\ref{fig:motivation}.
	Such feature alignment alleviates the parameter averaging conflicts and providing higher convergence accuracy.
	This also benefits the convergence speed as discussed in the next section.

\begin{figure*}[]
\centering
\vspace{-3mm}
\includegraphics[width=5.5in]{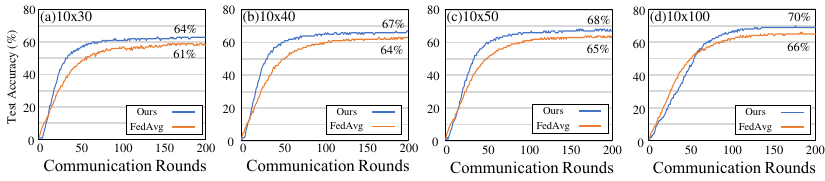}
\vspace{-5mm}
\caption{Convergence Speed Comparison between FedAvg and Proposed Method ({VGG16 on CIFAR100}).}
\label{fig:conv_speed}
\vspace{-5mm}
\end{figure*}

\vspace{-3mm}
\paragraph{Convergence Speed Improvement.}
%Besides the accuracy improvement, we demonstrate that our framework can also providing better convergence speed. 
	Four settings are selected similar with CIFAR10 non-IID spectrum settings, \textit{e.g.}, 10 nodes with each node having partial classes presented.
	Fig.~\ref{fig:conv_speed} shows the test accuracy curves of FedAvg and ours through the entire training process.
	% Our method consistently shows higher convergence accuracy (\textbf{+5\%$\sim$11\%} than FedAvg), as well as higher convergence speed.
	In all the non-IID cases, our method consistently shows higher convergence speed with only 50-80 rounds for the optimal performance, while FedAvg usually needs at least 100 rounds. 
	One exception is the 10x100 IID setting, the initial convergence of FedAvg seems to be faster, but our method soon exceeds it after 50 epochs and finally achieves \textbf{+4\%} accuracy than FedAvg. 
	% As previous work has mentioned~\cite{fed-noniid}, the parameter divergence issue is usually less severe under IID data distributions.
	% Therefore, our structural feature regulation has less advantages in terms  

\begin{table}[]
\vspace{-3mm}
\footnotesize
\parbox{.49\linewidth}{
\caption{Non-IID Spectrum (CIFAR10).}
\setlength{\tabcolsep}{1.7mm}{
\begin{tabular}{cccccc}
\toprule
                           & N * C & 10x3 & 10x4 & 10x5 & 10x10 \\ \midrule
\multirow{2}{*}{VGG9}      & FedAvg         & 82\% & 84\% & 85\% & 88\%  \\ 
                           & Ours           & 83\% & 88\% & 88\% & 90\%  \\ \midrule
\multirow{2}{*}{MbNet}   & FedAvg         & 67\% & 71\% & 79\% & 85\%  \\ 
                           & Ours           & 86\% & 88\% & 90\% & 91\%  \\ \bottomrule
\end{tabular}
\label{table:full_spectrum}}}
\hfill
\parbox{.49\linewidth}{
\caption{Node Scalability (CIFAR10).}
\setlength{\tabcolsep}{1.7mm}{
\begin{tabular}{cccccc}
\toprule
                           & N * C & 10x5 & 20x5 & 50x5 & 100x5 \\ \midrule
\multirow{2}{*}{VGG9}      & FedAvg         & 85\% & 86\% & 83\% & 83\%  \\
                           & Ours           & 88\% & 88\% & 86\% & 87\%  \\ \midrule
\multirow{2}{*}{MbNet} 	   & FedAvg         & 79\% & 85\% & 81\% & 78\%   \\
                           & Ours           & 90\% & 90\% & 89\% & 88\%   \\ \bottomrule
\end{tabular}}}
\normalsize
\end{table}

\begin{figure*}[!b]
% \centering
\vspace{-3mm}
\includegraphics[width=5.5in]{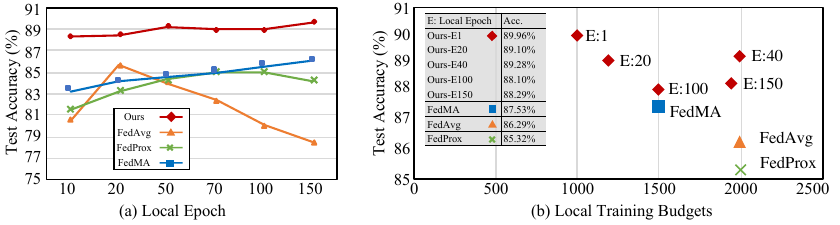}
\vspace{-6mm}
\caption{Local Epoch Influence and Local Training Budgets Comparison with SOTA.}
\label{fig:local_epoch}
\vspace{-3mm}
\end{figure*}

\vspace{-4mm}
\subsection{Performance Comparison with SOTA (FedProx \& FedMA)}
\vspace{-3mm}
In this part, we conduct performance comparison between our work with SOTA methods including FedProx and FedMA.
	We compare our performance in two aspects: (1) convergence accuracy under different local training epoch numbers $e$; (2) the optimal performance under given local training budgets.
	The overall results are shown in Fig.~\ref{fig:local_epoch}, where the SOTA results are referred from~\cite{fedma}. 
	All experiments use the same setting (VGG9, 16 nodes) with the dirichlet data distribution of CIFAR10.

\vspace{-3mm}
\paragraph{Communication Efficiency with Different Local Epochs.}
As previous work studied, with higher local training epochs, the frequency of the communication in FL could be effectively reduced.
	However, longer local training epochs can also lead to sub-optimal performance, since the model collaboration are less frequent with potential model divergence.
	Therefore, we aim to achieve high convergence accuracy with as longer local epochs as possible.
	Fig.~\ref{fig:local_epoch} (a) shows the convergence accuracy performance comparison under local epoch settings \{10, 20, 50, 70, 100, 150\}.
	All models are trained with 54 averaging rounds as per settings in~\cite{fedma} for fair comparison.
	Clearly, our framework shows the best performance under all settings, improving FedMA by \textbf{+3\%$\sim$5\%} accuracy.

\vspace{-3mm}
\paragraph{Computational Efficiency with Local Training Budgets.}
%Besides the communication overheads, FL also requires edge nodes to conduct local training frequently which consumes the local training budgets.
	%Therefore, we further evaluate the model performance under different local training budgets.
Specifically, the results of baseline FedAvg and SOTA works are their corresponding optimal performance reported in~\cite{fedma}.
	As we can see that, our method could achieve \textbf{+0.6\%$\sim$2.4\%} accuracy than FedMA with similar or less local training budgets.
	For FedAvg and FedProx, our method could achieve \textbf{+3.0\%} and \textbf{+4.0\%} accuracy improvement with the same training budget, respectively.

\vspace{-3mm}
\subsection{Ablation Study of Design Components}
\vspace{-3mm}
We study the performance influence of three design strategies in our framework, including the normalization layer influence, grouping number selection, as well as the sharing layer depth selection.

\vspace{-3mm}
\paragraph{Normalization Strategy Analysis.}

\begin{figure}[!tb]
\parbox{.49\linewidth}{
\centering
\vspace{-2mm}
\includegraphics[width=2.5in]{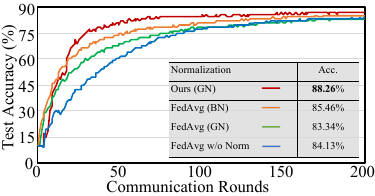}
\vspace{-1.5mm}
\caption{Normalization Strategy Analysis.}
\label{fig:batchnorm}}
\hfill
\parbox{.49\linewidth}{\centering
\vspace{-2mm}
\includegraphics[width=2.5in]{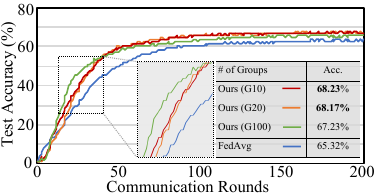}
\vspace{-1.5mm}
\caption{Number of Groups Analysis.}
\label{fig:groups}
}
\vspace{-5mm}
\end{figure}

For normalization strategy analysis, we compare our work and FedAvg under the data distribution of (VGG9, CIFAR10, 10x4 non-IID).
	The results are shown in Fig.~\ref{fig:batchnorm}.
	The baseline FedAvg without normalization achieves 84.13\% accuracy. 
	FedAvg+BN helps improve the accuracy to 85.46\%, while FedAvg+GN hurts the model performance, degrading the accuracy to 83.34\%.
	Utilizing the same GN settings, our work instead achieves the best accuracy 88.26\%. 
	This further demonstrates that our grouped structure helps alleviate the statistics divergence within groups, thus the group normalization could benefit the model performance.

\vspace{-3mm}
\paragraph{Grouping Number Analysis.}
In this part, the performance robustness of our model is demonstrated with complementary benefits under different group number selection.
	% while providing certain different convergence benefits.
	The overall results are shown in Fig.~\ref{fig:groups} (VGG16, CIFAR100, 10x50 non-IID).
Our 10-group and 20-group structures both achieve the optimal performance around $\sim$68\%, providing +2.7\% accuracy improvment than FedAvg -- 65.3\% with the original structure.
	The 100-group structure, though achieving sub-optimal accuracy improvement (+1.9\% than FedAvg), shows the highest convergence speed (the green curve).
	We hypothesize the convergence speedup is brought by its most fine-grained feature alignment benefits (1 class mapped to 1 group).
	But oppositely, with 100 groups, the per-path capacity can be limited. Though sub-optimal accuracy, it verifies the group size trade-off as discussed in Sec.~\ref{sec:3}.

\begin{figure}[!b]
\centering
\captionlistentry[table]{}
\captionsetup{labelformat=andtable}
\vspace{-3mm}
\caption{Grouping Depth Sensitivity. \textbf{Left:} Total Variance of each layer. \textbf{Right:} Performance under different grouping depth, showing our models' performance is robust to depth selection (10x5, 10x50 non-IID).}
\vspace{-2mm}
\parbox{.32\linewidth}{
% \begin{figure}[]
\includegraphics[width=2.2in]{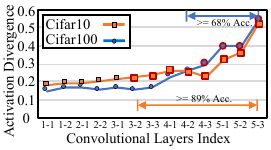}
\label{fig:depth}
% \end{center}
}
\qquad
\parbox{.6\linewidth}{
\centering
\vspace{-5mm}
\renewcommand\arraystretch{1.2}
\setlength{\tabcolsep}{1.0mm}{
\footnotesize
\hspace{8mm}
\begin{tabular}{c|cccccc|c}
\toprule
VGG16      & C1-2 & C2-2 & C3-2 & C4-2  & C5-2  & FC  & None \\ \midrule
CIFAR10  & 82\%    & 87\%    & \textbf{89\%}  & \textbf{90\%} & \textbf{89\%}          & 87\% & 86\%     \\ \hline
CIFAR100 & 53\%    & 62\%    & 67\%  & \textbf{68\%}   & \textbf{69\%} & \textbf{68\%} & 65\%    \\ \bottomrule
\end{tabular}
\label{tab:depth}}}
\vspace{-5mm}
\end{figure}

\vspace{-3mm}
\paragraph{Grouping Depth Analysis.}
We also demonstrate that our framework's performance is robust to the grouping depth hyper-parameter selection in Figure~\ref{tab:depth}.
	Specifically, our model shows nearly optimal performance 89\%$\sim$90\% in large ranges of depth selection, \textit{e.g.}, from C3-2 to C5-2 on CIFAR10.
	Also, the total variance of the model (a pre-trained model with minimum 50 epoch pre-training) offers good indication of the layer-wise feature divergence.

\normalsize

% \begin{figure}[]
% \centering
% \vspace{-2mm}
% \includegraphics[width=2.5in]{granularity}
% \vspace{-2mm}
% \caption{Path Granularity: one-to-one and one-to-many mapping.}
% \label{fig:path}
% \vspace{-5mm}
% \end{figure}
\vspace{-2mm}
\section{Conclusion}
\label{sec:conc}
\vspace{-3mm}
In this paper, we proposed a novel federated learning framework to resolve FL models' chaotic fusion problem by establishing a firm structure-information alignment.
	In the process of proposing this method, we adopted feature-oriented parameter interpretation and proposed a novel feature-oriented regulation method (\textit{$\Psi$-Net}) to ensure explicit feature information allocation in different neural network structures.
	Such a regulation method adjusts the local model architecture according to their data and task distribution at the very early training stage, and continuously maintain not only structure but also information alignment for better FL performance.
	To our best knowledge, this is the very first work that solves the FL alignment problem directly from a model regulation perspective, which significantly improves the FL performance and applicability in various heterogeneous settings.
	%Experiment demonstrates significant improvement in convergence speed, accuracy and computation/communication cost, outperforming previous state-of-the-art works by large margins.

{\small
\bibliographystyle{ieee_fullname}
\bibliography{dac_shuffled}
}

\end{document}